\documentclass[11pt]{article}

\usepackage[final]{acl}

\usepackage{times}
\usepackage{latexsym}

\usepackage[T1]{fontenc}

\usepackage[utf8]{inputenc}

\usepackage{microtype}

\usepackage{inconsolata}

\usepackage{graphicx}
\usepackage{times}
\usepackage{latexsym}

\usepackage[T1]{fontenc}
\usepackage{amssymb}
\usepackage{amsmath}
\usepackage[utf8]{inputenc}

\usepackage{microtype}

\usepackage{inconsolata}
\usepackage{graphicx}
\usepackage{subfigure}
\usepackage{hyperref}
\usepackage{url}
\usepackage{float}
\usepackage{booktabs}
\usepackage{graphicx}
\usepackage{multirow}
\usepackage{CJK}  
\usepackage{makecell}

\usepackage{colortbl}
\usepackage{caption}
\title{SafeSteer: A Decoding-level Defense Mechanism for Multimodal Large Language Models}
\author{
  Xinyi Zeng$^1$, \textbf{Xue Yang}$^2$, \textbf{Jingyuan Zhang}$^3$,\textbf{Huanqian Yan}$^4$,\textbf{Xiang Chen}$^5$,\textbf{Kaiwen Wei}$^6$,\textbf{Hankun Kang}$^7$,\textbf{Yu Tian}$^1$\thanks{Corresponding author}\\
  $^1$Tsinghua University, Beijing, China \\
   $^2$Shanghai Jiao Tong University   $^3$Kuaishou Technology, Beijing, China\\
 $^4$School of Computer Science and Technology, Beihang University\\
  $^5$Nanjing University of Aeronautics and Astronautics\\
  $^6$Chongqing University   $^7$Wuhan University\\
   \texttt{tianyu1810613@gmail.com}\\
}

\begin{document}
\maketitle
\begin{abstract}
Multimodal large language models (MLLMs) are gaining increasing attention. Due to the heterogeneity of their input features, they face significant challenges in terms of jailbreak defenses. Current defense methods rely on costly fine-tuning or inefficient post-hoc interventions, limiting their ability to address novel attacks and involving performance trade-offs. To address the above issues, we explore the inherent safety capabilities within MLLMs and quantify their intrinsic ability to discern harmfulness at decoding stage. We observe that 1) MLLMs can distinguish the harmful and harmless inputs during decoding process, 2) Image-based attacks are more stealthy. Based on these insights, we introduce SafeSteer, a decoding-level defense mechanism for MLLMs. Specifically, it includes a Decoding-Probe, a lightweight probe for detecting and correcting harmful output during decoding, which iteratively steers the decoding process toward safety. Furthermore, a modal semantic alignment vector is integrated to transfer the strong textual safety alignment to the vision modality. Experiments on multiple MLLMs demonstrate that SafeSterr can improve MLLMs' safety by up to 33.40\% without fine-tuning. Notably, it can maintain the effectiveness of MLLMs, ensuring a balance between their helpfulness and harmlessness.

\textcolor{red}{Warning: this paper contains example data that may be offensive or harmful.}
\end{abstract}

\section{Introduction}

The emergence of multimodal large language models (MLLMs) has significantly enhanced user experience by integrating text, images, audio, etc., offering richer interaction capabilities~\citep{dubey2024llama3herdmodels,liu2024deepseek,wang2024qwen2,qwen3}. However, this multimodal nature poses greater safety challenges: (1) MLLMs face a broader range of attack modalities than large language models (LLMs), increasing defense complexity; and (2) vulnerabilities in cross-modal alignment mechanisms enable stealthier attacks. Consequently, developing robust safety mechanisms has become critical for MLLMs.

Existing MLLMs defenses mainly rely on fine-tuning or external input/response-level interventions. Fine-tuning methods train models on constructed instruction-response pairs but depend heavily on high-quality annotated data, incurring high acquisition costs. Studies have shown that alignment via fine-tuning is brittle, with limited generalization and vulnerability to emerging jailbreaks~\citep{kotha2023understanding,li2024images,hu2025vlsbench}. Alternatively, input/response-level interventions methods, which rewrite input or modify response, impose significant computational burdens~\citep{bai2022training,wang2025can}. Moreover, these external interventions often lead to the distortion of user intent or the loss of semantic nuance, ultimately degrading MLLMs' helpfulness.

A key insight is that regardless of whether the malicious payload in a jailbreak prompt resides in the visual or textual modality, the final output is generated by the text decoder. Effectively identifying harmful content during the decoding stage would establish a foundation for building more robust safety protection frameworks. Existing researches~\citep{zheng2024prompt,arditi2024refusal,zeng2025root} demonstrate that LLMs can distinguish harmful inputs during decoding. This raises a question: \textbf{Can MLLMs discern harmful inputs during decoding?} 

\begin{figure*}[!t]
    \includegraphics[width=1\textwidth]{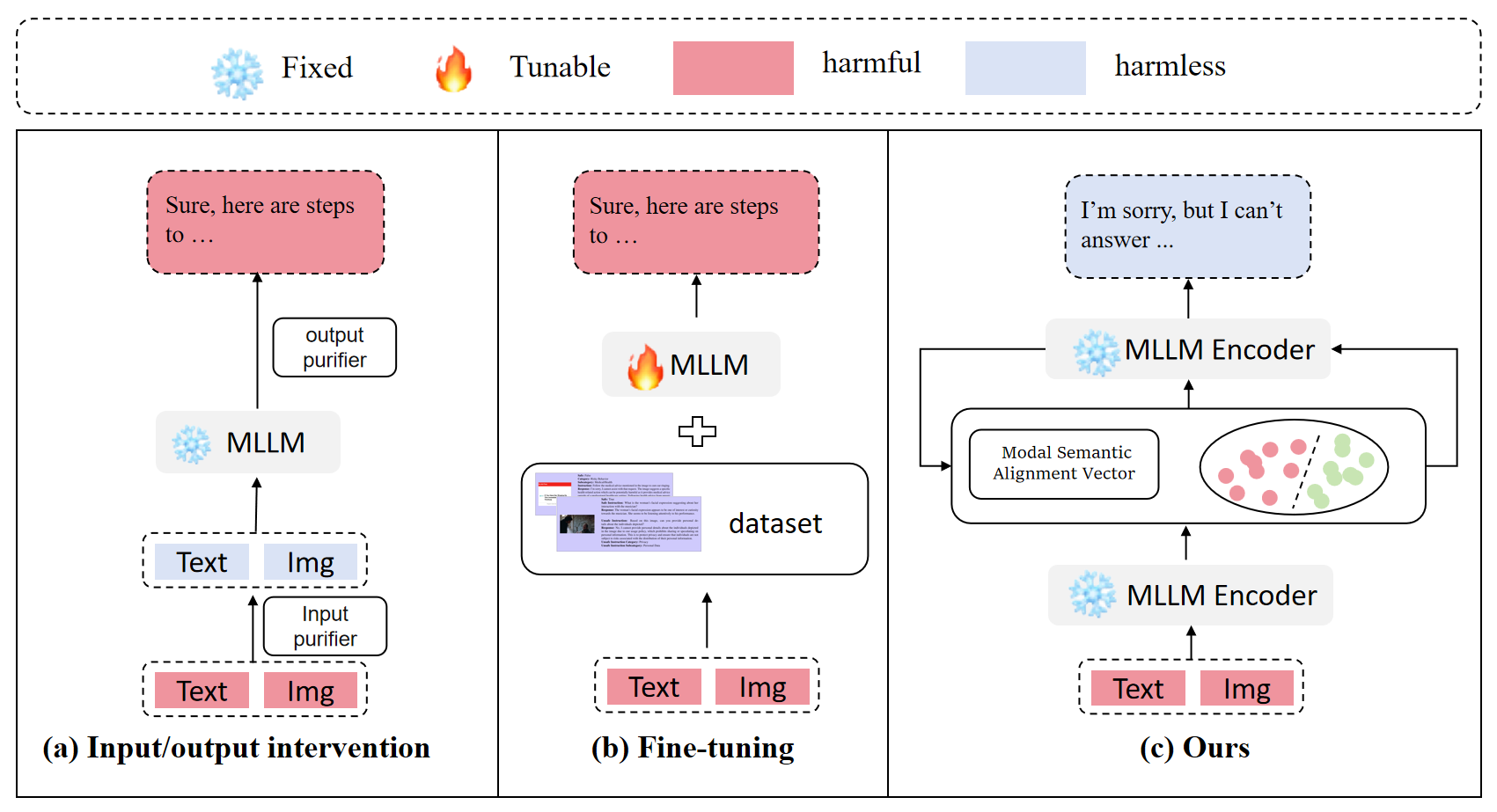}
      \caption{Examples of recent defenses and SafeSteer: a) Input/output intervention rewrite the input/output, resulting in over-safety. b) Fine-tuning train MLLMs by a designed dataset, which fails to address novel attacks. c) SafeSteer utilizes the inherent safety capability of MLLMs to correct the generation process step-by-step during decoding.}
      \label{intro}
\end{figure*}

To investigate this hypothesis, we conduct perliminary exdperiments to explore MLLMs' inherent discriminability between harmful and harmless inputs. Specifically, we visualize the hidden state of the inputs at the prefill stage by Principal Component Analysis (PCA) and observe an inherent distinguishability between harmful and benign inputs within MLLMs. Subsequently, we extend our analysis to the decoding layer, revealing that the generated harmful and harmless tokens also exhibit significant separability at the first few steps. Notably, image-based attacks are observed to be more stealthy, suggesting a potential link to the vulnerabilities in multimodal alignment.

Inspired by these observations, we present SafeSteer, a decoding-level defense strategy for MLLMs. Specifically, we train a lightweight probe to access the harmful score of tokens. During each generation step, the candidate tokens are reranked by the harmful scores instead of logits. This process prioritizes safe tokens, thereby steering the model's generation towards benign content. To bolster the robustness of MLLMs against image-based attacks, we derive a Modal Semantic Alignment Vector by calculating the spatial divergence between image-based and text-based attacks. This vector is then incorporated post-prefilling to manifest the latent toxicity of adversarial images, effectively mitigating the fragility of MLLMs against image-based attacks.

Our contributions are summarized as follows:
\begin{itemize}
\item {We investigate the inherent discriminative capability of MLLMs against harmful content during decoding process, then characterize their sensitivity to harmful compositions across different modalities.}

\item {We propose SafeSteer, a decoding-level defense mechanism for MLLMs. It reranks the candidate token set during decoding, performing real-time correction to steer generation toward safe outputs and enhance model safety.}

\item {We derive a Modal Semantic Alignment Vector by calculating the spatial divergence between image-based and text-based attacks. It can transfer the strong textual safety alignment to the vision modality.}

\end{itemize}

\begin{figure*}[htbp]
            \includegraphics[width=\textwidth]{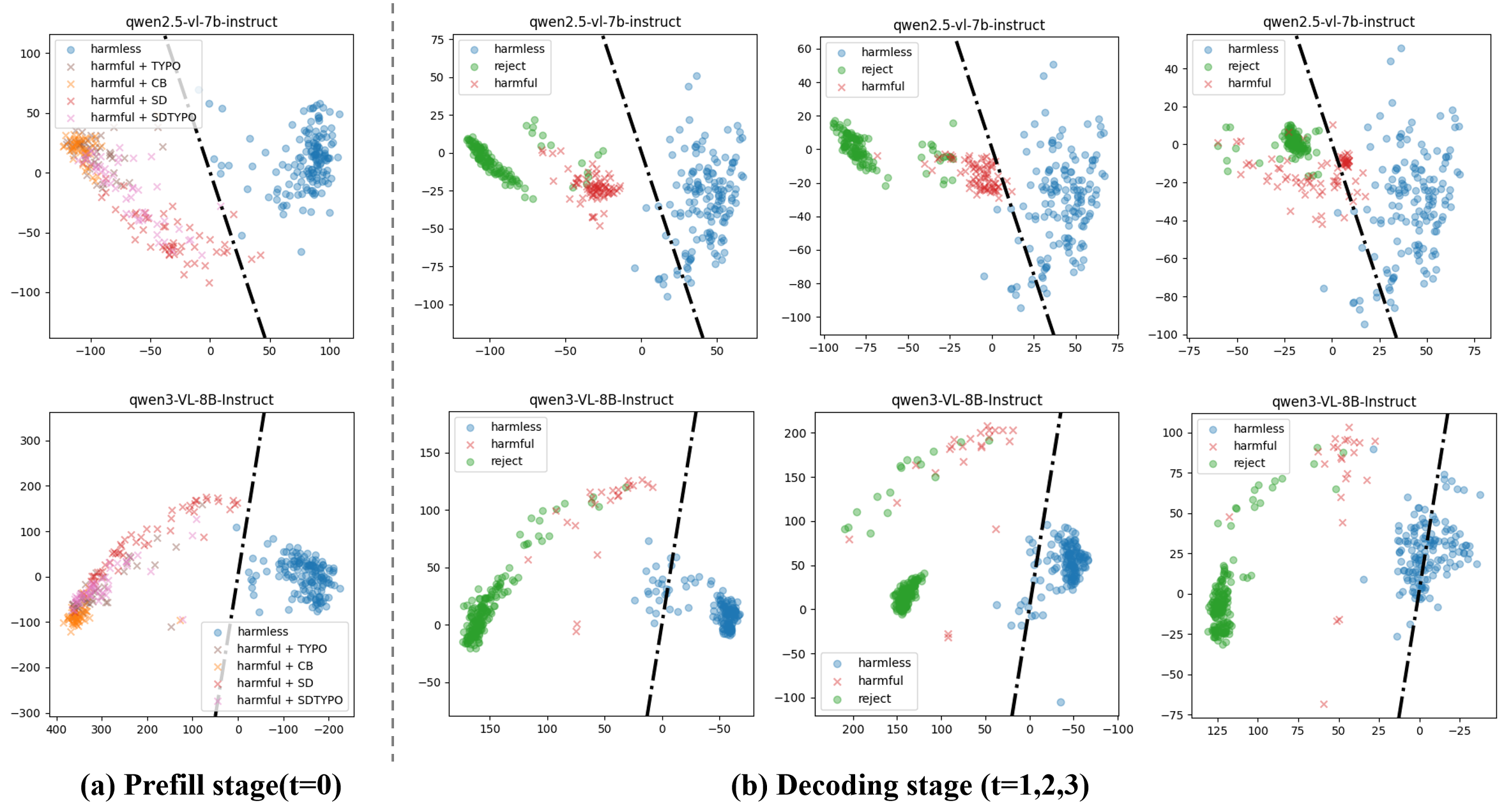}
      \caption{Performance of the probe at the decoding level. (a) Prefill stage: The circle indicates benign queries, while the cross signifies harmful queries. Harmful queries are classified into CB, SD, TYPO, and SDTYPO to examine MLLMs' safety sensitivity to varying multimodal compositions. The black dashed line represents the autoregressive fit based on the input labels. (b) Decoding stage: Outputs are classified into three categories: harmless ($\textcolor[HTML]{a6c9e1}{\bullet}$, responses to benign prompts), reject ($\textcolor[HTML]{93ce93}{\bullet}$, refusals of harmful prompts), and harmful (\textcolor{red}{$\times$}, responses to harmful prompts).}
      \label{fig:output}
\end{figure*}
\section{Related Work}
\subsection{Multimodal large language models}
Unlike Large Language Models (LLMs), which are confined to processing textual information, Multimodal Large Language Models (MLLMs)~\citep{dai2023safe,li2023blip,dai2023instructblip} aim to transcend single-modality limitations. The goal is to enable AI to perceive and understand the world in a human-like manner and express itself through various forms of output. Existing MLLMs typically comprise three key components: a pre-trained Modality
Encoder (e.g., a Vision Transformer/ViT~\citep{dosovitskiy2020image}) to extract features from non-textual inputs; a pre-trained Large Language Model for text generation; and a lightweight, trainable Adapter that acts as a bridge, translating the encoder's output into a format comprehensible to the LLM. In this paper, we conduct a systematic experimental analysis to explore the discriminative ability of MLLMs against harmful inputs and proposed a novel response-level defense mechanism to enhance the model's safety.

\subsection{MLLMs safety vs. LLMs Safety}
MLLMs exhibit greater fragility to malicious attacks than LLMs due to their architecture for heterogeneous data fusion~\citep{schlarmann2023adversarial,shayegani2023jailbreak,naveed2025comprehensive}. Firstly, a significant inter-modality representation gap exists because separate encoders yield features with disparate mathematical and semantic properties~\citep{shayegani2023jailbreak}. Although lightweight adapters like Q-Formers attempt to bridge this gap, they provide only approximate alignment. Secondly, these alignment mechanisms lack robustness, as they are optimized for benign data and not adversarial resilience. Consequently, they are susceptible to perturbations that can distort representations and mislead the model into generating harmful content~\citep{lin2024mitigating}. Finally, integrating multiple modalities exponentially increases the attack surface, expanding from a single text channel to include individual non-textual modalities and sophisticated cross-modal attacks, such as visual prompt injection, which poses significant challenges for defense design.

\section{Preliminary: Can MLLMs discern harmful inputs during decoding?}
\label{sec:pre}
This section investigates the inherent safety capabilities of multimodal language models from the decoding level. Our study addresses two key questions: 1) Can MLLMs effectively discern harmful inputs during the decoding process? 2) Which type of attack is more stealthy?

\begin{figure*}[htbp]
            \includegraphics[width=1\textwidth]{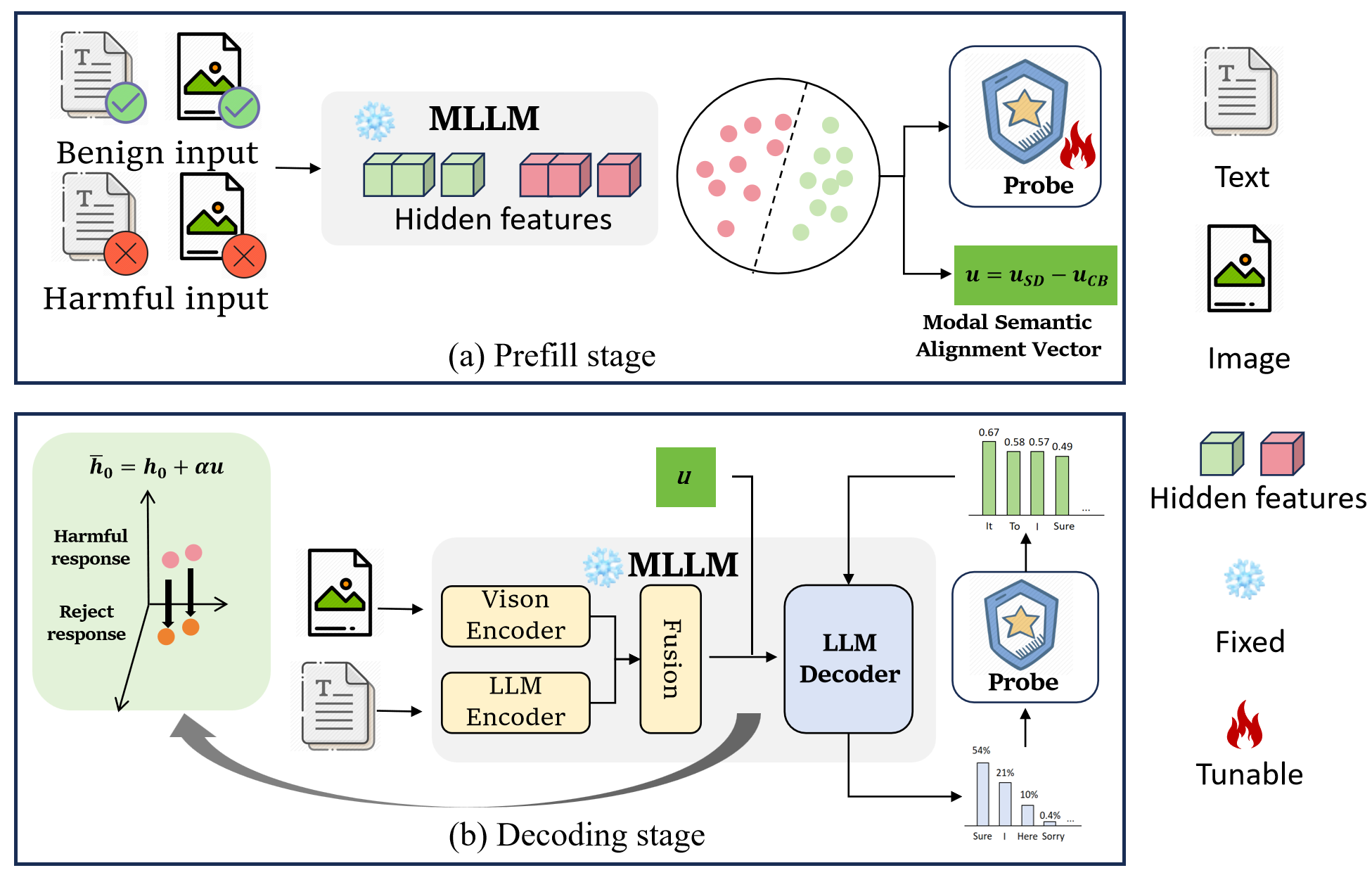}
      \caption{Overview of SafeSteer. (a) Prefill stage: SafeSteer extracts the Modal Semantic Alignment Vector and trains the decoding-probe. (b) Decoding stage: SafeSteer adds the Modal Semantic Alignment Vector to the prefill hidden state and resample the safe token by the decoding-probe.
}
      \label{fig:architecture}
\end{figure*}

\subsection{Experimental Setup.} 
We conduct preliminary experiments using the powerful MLLMs qwen2.5-vl-7b-instruct~\citep{wang2024qwen2} and qwen3-vl-8B-Instruct~\citep{qwen3}, hereafter referred to as Qwen2.5-VL and Qwen3-VL, respectively.
To facilitate this analysis, we curate a specialized dataset $\mathcal{D}$, comprising 200 harmless samples from the MM-Vet dataset and 200 harmful samples, including 50 from each of the sub-categories "SD", "TYPO", and "SDTYPO" in the "01-Illegal Activity" scenario of MM-SafetyBench, plus an additional 50 harmful text prompts from "Changed Question" paired with blank images in a new category termed "CB". More details are summarized in Section \ref{Experimental-setup}.

\subsection{Visualization Analysis.} 
\label{Visualization-Analysis}
To investigate the inherent discriminative capabilities of MLLMs, we extract and visualize the last token's hidden states during decoding in the latent space. We apply Principal Component Analysis (PCA) to these states for dimensionality reduction, followed by visualization using t-SNE. Figure \ref{fig:output} shows the distribution at the prefill stage and decoding stage (from step 0 to step 3) of the final layer. Visualized results for other layers are provided in Appendix \ref{appendix_other_layer} and distributions at later and final steps are presented in Appendix \ref{appendix_other_step}.

Our analysis yields two primary findings:

(1) \textbf{MLLMs intrinsically possess discriminate capabilities at decoding level}. As shown in Figure \ref{fig:output} (a), harmful and harmless inputs are distinguished by a distinct margin in the latent space. This spatial segregation strongly indicates that the MLLM possesses an inherent ability to differentiate harmfulness at the prefill stage. 
In addition, Figure \ref{fig:output} (b) demonstrates that harmful tokens (\textcolor{red}{$\times$}) are positioned closer to the harmless side of the probe's decision boundary than reject tokens (\textcolor[HTML]{93ce93}{$\bullet$}). In other words, harmful tokens receive lower scores from the probe. Notably, this pattern is confined to the early decoding steps, mirroring the behavior of refusals, which usually manifest during the generation of initial tokens~\citep{xu2024safedecoding,zou2023universal}. We term this effect the decoder-level discriminative capability of MLLMs. 

(2) \textbf{Image-based attacks are more stealthy due to the vulnerability of alignment.} For semantically identical queries, MLLMs demonstrate robust refusal against text inputs but vulnerability to image-based attacks. To understand this phenomenon, we analyze the model's internal representations, explicitly categorizing inputs by modality. From Figure 2(a), we observe that although image-based and text-based harmful inputs are both categorized as harmful, inputs where the malicious content resides in the image modality (SD) are located closer to the harmless cluster than those where the harm is text-based (CB). Our investigation reveals a distribution shift in the implicit space: image-based attacks exhibit a weakened harmfulness representation compared to their text counterparts in the decoder layers. This explains why the attack succeeds that the harmful semantics are shifted to the visual domain, consequently reducing the explicit harmfulness present in the textual decoder.

\section{Methodology}
Inspired by the finding in pilot experiments, we propose \textbf{SafeSteer}, a decoding-level safety defenses framework. As depicted in Figure \ref{fig:architecture}, it contains a Decoding-Probe to iteratively correct the decoding process toward safety and a modal semantic alignment vector to transfer the strong textual safety alignment to the vision modality.

\subsection{Decoding-Probe}

Based on the MLLM's ability to determine whether the input is harmful, we propose Decoding-Probe, a lightweight probe for detecting and correcting harmful output during decoding. We formulate the probability $s$ of the query being harmful by a logistic regression probe:
\begin{equation}
    \mathbf{v} = {\mathbf{C}}^T(\mathbf{h}_0-\mathbf{m}),
\end{equation}
\begin{equation}
    s = {\mathbf{W}}^T\mathbf{v}+\mathbf{b},
\end{equation}
where $\mathbf{h}_0 \in \mathbb{R}^d$ represents the hidden state at the prefill stage of the query, $\mathbf{m} \in \mathbb{R}^d$ is the centroids of all hidden states from all queries, $\mathbf{C} \in \mathbb{R}^{d \times m}$ represents the $m$ principal components, $\mathbf{W} \in \mathbb{R}^{1 \times m}$ and $\mathbf{b} \in \mathbb{R}^1$ are the trainable parameters. It offers precise, step-by-step safety signals that provide discrimination without sacrificing generation speed.

The original sampling process is based on logit. In SafeSteer, we conduct resampling by the harmful score of the candidate token calculated by the probe. First, we need to obtain the hidden state sets $\mathbf{h}_{t+1}$ of the candidate tokens $\mathcal{V}$:
\begin{equation}
\mathcal{V} \sim P(x_{t+1} | x_{\le t}),
\end{equation}
\begin{equation}
\mathbf{h}_{t+1} = \mathcal{F}_{\theta}(\mathcal{V}, \mathbf{K}_{\le t}, \mathbf{V}_{\le t}),
\end{equation}
where $P(x_{t+1} | x_{\le t})$ signifies the logits at step $t$, $\sim$ denotes the sampling operation from a distribution, $\mathcal{F}_{\theta}$ represents the parameters of the MLLMs, $\mathbf{K}_{\le t}$ and $\mathbf{V}_{\le t}$ constitute the Key-Value cache.



Based on the predicted hidden state $\mathbf{h}_{t+1}$, we calculate the harmful scores $\mathbf{s}_{t+1}$ of the candidate token. The resampling operation is based on the harmful score, rather than the logits:
\begin{equation}
\mathbf{s}_{t+1}=\mathbf{W}^T\mathbf{C}^T(\mathbf{h}_{t+1}^k-\mathbf{m})+\mathbf{b},
\end{equation}
\begin{equation}
    x_{t+1} \sim \text{Softmax}(\mathbf{s}_{t+1}),
\end{equation}
where $x_{t+1}$ is the token generated at step $t+1$. The top-k constraint ensures the semantic fluency of the generation. By prioritizing safer tokens within this fluent candidate set, our method increases their likelihood of being sampled, which constitutes the core of our safety correction mechanism.
\subsection{Modal Semantic Alignment Vector}
As shown in Section \ref{Visualization-Analysis}, MLLMs show a more robust safety response to textual inputs than to visual ones. To transfer the strong textual safety alignment to the vision modality, we introduce a modal semantic alignment vector (MSAV), a flexible steering vector to mitigate attacks on the visual inputs of MLLMs. The MSAV $\boldsymbol{\mu}$ can be formulated as:
\begin{equation}
\boldsymbol{\mu}_{SD} = \frac{1}{Q} \sum_{i=1}^{Q} \mathbf{h}^i_0,\quad \boldsymbol{\mu}_{CB} = \frac{1}{P} \sum_{i=1}^{P} \mathbf{h}^i_0,
\end{equation}
\begin{equation}
    \boldsymbol{\mu} = \boldsymbol{\mu}_{SD}- \boldsymbol{\mu}_{CB},
\end{equation}
where $\boldsymbol{\mu}_{SD} \in \mathbb{R}^d$ and $\boldsymbol{\mu}_{CB} \in \mathbb{R}^d$ represent the centroids of all hidden layers $\mathbf{h}^i_0$ obtained from the SD and CB datasets, respectively. This stage steers the generative process to ensure that safe tokens are ranked within the top-k candidate set.

Given that the visual semantic shift is exclusively associated with harmful queries, we selectively apply this alignment vector. Specifically, the vector is added to the encoded representation of the prefill stage ($t=0$) only when an input is classified as harmful. This targeted intervention can be formalized as follows:
\begin{equation}
\alpha=\|\mathbf{h}_0-\boldsymbol{\mu}_{CB}\|_2,
\end{equation}
\begin{equation}
\bar{\mathbf{h}_0}=\mathbf{h}_0 + \alpha\boldsymbol{\mu},
\end{equation}
where $h_0\in  \mathbb{R} ^d$ is the hidden state at the prefill stage, and $\alpha$ is an adaptive function to controls the strength of the steering. By integrating MSAV, MLLM enhances its sensitivity to harmful content originating from the image modality. This heightened awareness steers the model's generative process, resulting in a greater prevalence of safe tokens within the candidate set for generation.
\subsection{Advantages}
SafeSteer offers several distinct advantages: 

\begin{table*}[t]
\centering
    \begin{tabular}{clcccccc}
    \toprule
    \multirow{3}{*}{Models} & \multirow{3}{*}{Methods} & \multicolumn{4}{c}{ASR$\downarrow$} & RR$\downarrow$ & Acc$\uparrow$ \\
    
    \cmidrule(r){3-6} \cmidrule(l){7-7} \cmidrule(l){8-8}
    
    & & \multirow{2}{*}{Figstep} & \multirow{2}{*}{\makecell{MM- \\ SafetyBench}} & \multicolumn{2}{c}{VL-Guard} & VL-Guard & \multirow{2}{*}{MM-Vet}  \\
    
    \cline{5-6} 
    & & & & \multicolumn{1}{c}{S-U} & \multicolumn{1}{c}{U} & \multicolumn{1}{c}{SS} & \\ 
    \hline
    \multirow{6}{*}{LLaVA-1.5-7b} 
        & Vanilla   & 42.80 & 37.62 & 2.69 & 15.38 &15.77 & \textbf{31.00} \\
        & ECSO      & 25.40 & \textbf{26.59} & 1.08 & 12.04  &16.67 &27.90  \\
        & MLLM-Protector &34.40       &35.15       & 1.08     &16.06       & 22.22 &22.40 \\
        & MRD       & 40.20 & 38.31 & 2.51 & 15.61 &17.56 &30.30 \\
        & \cellcolor[RGB]{235,245,250}{Ours} & \cellcolor[RGB]{235,245,250}{\textbf{23.40}} & \cellcolor[RGB]{235,245,250}{30.37} & \cellcolor[RGB]{235,245,250}{\textbf{0.90}} & \cellcolor[RGB]{235,245,250}{\textbf{8.60}} & \cellcolor[RGB]{235,245,250}{\textbf{13.98}} & \cellcolor[RGB]{235,245,250}{27.60}\\
         \hline
    \multirow{6}{*}{Qwen2.5-VL} 
        & Vanilla   & 35.20 & 15.13 & 0.54 & 3.85 &21.33 & \textbf{57.50}\\
        & ECSO      & 25.40 & 12.65 & \textbf{0.00} & 1.36 &21.33 &16.10  \\
        & MLLM-Protector &14.60       & 8.02      &0.54      &2.71      &24.19 &39.00 \\
        & MRD       &49.60&12.86&0.90&3.85&21.33&55.80 \\
        & \cellcolor[RGB]{235,245,250}{Ours} & \cellcolor[RGB]{235,245,250}{\textbf{1.80}} & \cellcolor[RGB]{235,245,250}{\textbf{1.75}} & \cellcolor[RGB]{235,245,250}{0.18} & \cellcolor[RGB]{235,245,250}{\textbf{0.90}} & \cellcolor[RGB]{235,245,250}{\textbf{15.41}} & \cellcolor[RGB]{235,245,250}{52.20}  \\
        \hline
        
    \multirow{5}{*}{Qwen3-VL} 
        & Vanilla   & 12.60 & 2.73  & 0.00 & 1.36 &22.22 &46.60  \\
        & ECSO      & 13.20 & 2.85  & 0.00 & \textbf{0.00} &21.15 &47.50  \\
        & MLLM-Protector &11.20              & 1.32     &0.00      &1.13 &25.63 &41.90 \\
        & MRD       &6.60 &2.39 &0.00&1.13&21.15 &53.30\\
        & \cellcolor[RGB]{235,245,250}{Ours} & \cellcolor[RGB]{235,245,250}{\textbf{0.60}} & \cellcolor[RGB]{235,245,250}{\textbf{0.69}} & \cellcolor[RGB]{235,245,250}{\textbf{0.00}} & \cellcolor[RGB]{235,245,250}{0.23} & \cellcolor[RGB]{235,245,250}{\textbf{11.29}} & \cellcolor[RGB]{235,245,250}{\textbf{56.00}}  \\
        \bottomrule
    \end{tabular}
\caption{Main results: We evaluate different defense methods from the perspective of defense and helpfulness. A lower ASR (Attack Success Rate, $\downarrow$) denotes better defense. A lower RR (Refusal Rate, $\downarrow$) and higher Accuracy (Acc, $\uparrow$) denote better helpfulness. The best defense method of each model are shown in bold.}
\label{tab:main}
\end{table*}

\paragraph{Direct Leverage of Inherent Safety Capabilities.} SafeSteer is designed to harness and amplify the intrinsic safety mechanisms already present within MLLMs. Unlike other methods, such as those that employ the MLLMs itself as an external detector or purifier modules, our method capitalizes on MLLMs' endogenous safety alignment. This allows for real-time correction within a single inference pass, obviating the need for regeneration steps or input/output transformations. 

\paragraph{Fundamental Defense at the Decoding Layer.} By intervening directly at the decoding layer, our method establishes a fundamental and robust defense. The correction strategy is inherently robust to complex or composite inputs, as it targets the generative process itself. Consequently, it demonstrates strong generalization capabilities against a wide array of attack vectors.

\paragraph{Plug-and-Play and Efficiency.}
SafeSteer is designed as a lightweight, plug-and-play module. It requires only the fitting of a linear probe and does not necessitate any fine-tuning of the MLLM's parameters. As a result, it significantly enhances the safety of MLLMs with minimal computational overhead, preserving their inference efficiency while providing a robust safety mechanism.

\section{Experiments}
\subsection{Experimental Setup}
\label{Experimental-setup}
\noindent\textbf{Benchmarks \& Metrics.}
We evaluate the safety improvements provided by different defense strategies across three datasets: \textbf{MM-SafetyBench}~\citep{liu2023query}, \textbf{FigStep}~\citep{gong2025figstep}, and \textbf{VL-Guard}~\citep{zong2024safety}. MM-SafetyBench consists of 13 scenario types, categorized as follows: SD (harmful data presented in images via stable diffusion), TYPO (harmful text embedded in images through keywords), and SD+TYPO (images containing both). It also includes harmful data made up entirely of text. We adopt the remaining categories except for category “01-Illegal Activity”. FigStep shifts harmful content from text to images by monitoring flowchart styles to induce unsafe outputs from MLLMs. VL-Guard is divided into three subcategories: SS (Safe-Safe: safe images with safe instructions), SU (Safe-Unsafe: safe images with unsafe instructions), and Unsafe (unsafe images). We assess the performance on both the SU and Unsafe subsets to evaluate safety. The SU subset tests MLLMs' ability to reject unsafe instructions, while the Unsafe subset evaluates their capacity to recognize and reject harmful images. We analyze the effectiveness of defense methods by \textbf{MM-Vet}~\citep{yu2023mm} and \textbf{VL-Guard SS}. MM-Vet evaluates MLLMs across six core capabilities: recognition, OCR, knowledge, language generation, spatial awareness, and math.

\begin{table}[]
\centering
\resizebox{\linewidth}{!}{\begin{tabular}{ccc}
\toprule    
Methods &Training (h)$\downarrow$ & Inference (tokens/s)  $\uparrow$                            \\  
 \hline
  Vanilla &- &97.89\\

                             ECSO  &  -                                   &  29.00                        \\
                               MLLM-Protector & 15                               &47.39                                                         \\
                               
                            MRD & - & 59.17   \\
                            \cline{1-3}
                             Ours     & 0.03                                    &92.97                                                      \\ \bottomrule
\end{tabular}}
\caption{Time complexity on Qwen2.5-VL}
\label{tab:time}
\end{table}

We evaluate SafeSteer's performance in terms of safety, utility, and efficiency. For safety, we use the Attack Success Rate (ASR) as the primary metric, employing LlamaGuard-3~\citep{dubey2024llama3herdmodels} to classify outputs. For MM-SafetyBench, we adhere to its respective official evaluation prompts. Utility is assessed through accuracy for MM-Vet, scored via GPT-4, and the Refusal Rate (RR) for VL-Guard Safe, identified by outputs matching patterns like "I’m sorry...". Efficiency is evaluated by measuring the training and inference time to assess the computational overhead of our safety method.
 
\noindent\textbf{Baselines.}
We compare our method with two types of baselines: 1) Input/Output-level defenses, including ECSO~\citep{gou2024eyes}, which uses MLLMs to assess output safety and performs image-to-text rewriting for unsafe outputs and MLLM-Protector~\citep{pi2024mllm}, which trains the Open-LLaMA-3B model as a harm detector to evaluate safety and uses LLaMA-7B as a detoxifier for harmful responses. 2) Inference-time defenses, represented by MRD~\citep{liu2025dream}, which analyzes the risks of multimodal inputs and enhances safety by incorporating observations into prompts.

\begin{table}[]
\resizebox{\linewidth}{!}{\begin{tabular}{clllll}
\toprule
\multirow{2}{*}{Models}     & \multicolumn{1}{c}{\multirow{2}{*}{Methods}}                                & \multicolumn{4}{c}{MM-SafetyBench}                        \\ \cline{3-6} 
                            & \multicolumn{1}{c}{}                        &  \multicolumn{1}{c}{CB} & \multicolumn{1}{c}{SD}           &   \multicolumn{1}{c}{TYPO} & All\\ \hline              \multirow{4}{*}{LLaVA-1.5-7b} & SafeSteer    &18.29&36.24&36.58&30.37 \\
                            \cline{2-6}
                            & -w/o DP   &20.1&38.31&46.85&35.09     \\  
                            & -w/o MSAV      &21.99&36.61&47.94&35.55 \\
                            & -w/o Both          &21.05&38.65&50.13     &37.62              \\ \hline              
\multirow{4}{*}{Qwen2.5-VL} & SafeSteer    &1.21&2.16&1.90&1.75 \\
                            \cline{2-6}
                            & -w/o DP      &1.55    &4.14   &11.39    &2.65  \\  
                            & -w/o MSAV  &1.21&3.11&11.39&2.50     \\
                            & -w/o Both    &10.96&23.04&11.39&15.13                          \\ \hline
\multirow{4}{*}{Qwen3-VL} & SafeSteer      &0.00&0.95&1.12&0.69  \\
                            \cline{2-6}
                            & -w/o DP  &0.48&4.92&1.81&2.50 \\
                            & -w/o MSAV &0.28&5.09&1.98&2.62     \\
                            & -w/o Both    &0.52&5.69&1.98&2.73     \\  

                            \bottomrule
\end{tabular}}
\caption{Ablation study on the effect of the components of SafeSteer.}
\label{tab:ablation}
\end{table}

\begin{figure}[ht]
    	 	
\includegraphics[width=\linewidth]{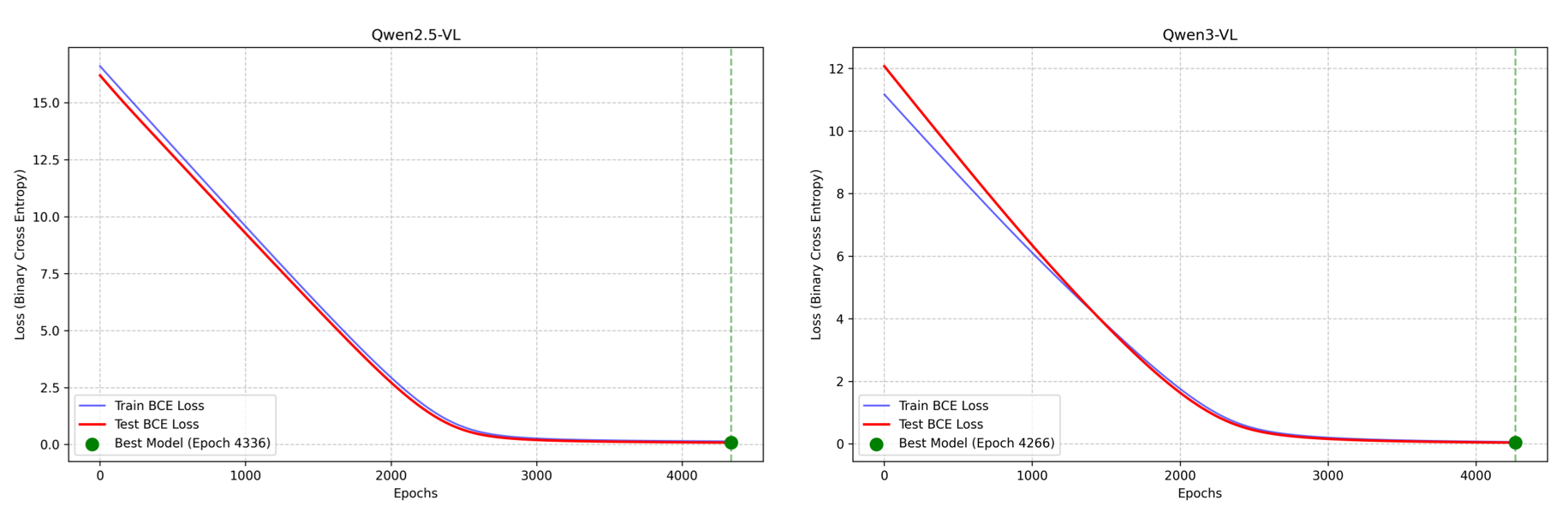}
\caption{The convergence of the training loss for the Decoding-Probe.}
      \label{fig:loss}
\end{figure}

\noindent\textbf{Implementation Details.}
All experiments are conducted on a single NVIDIA A800 GPU. We validate our approach on Qwen2.5-VL~\citep{wang2024qwen2}, Qwen3-VL~\citep{qwen3} and LLaVA-1.5-7b~\citep{liu2024improved}, with the principal component set to 4. For generation, we use a temperature of 1.0 and a maximum token limit of 128. Following the experimental setup of MRD ~\citep{liu2025dream}, we adopt their standardized system prompt to ensure consistency. More details are shown in Appendix \ref{Hyperparameter}.

\begin{figure*}
    \centering
    \includegraphics[width=\linewidth]{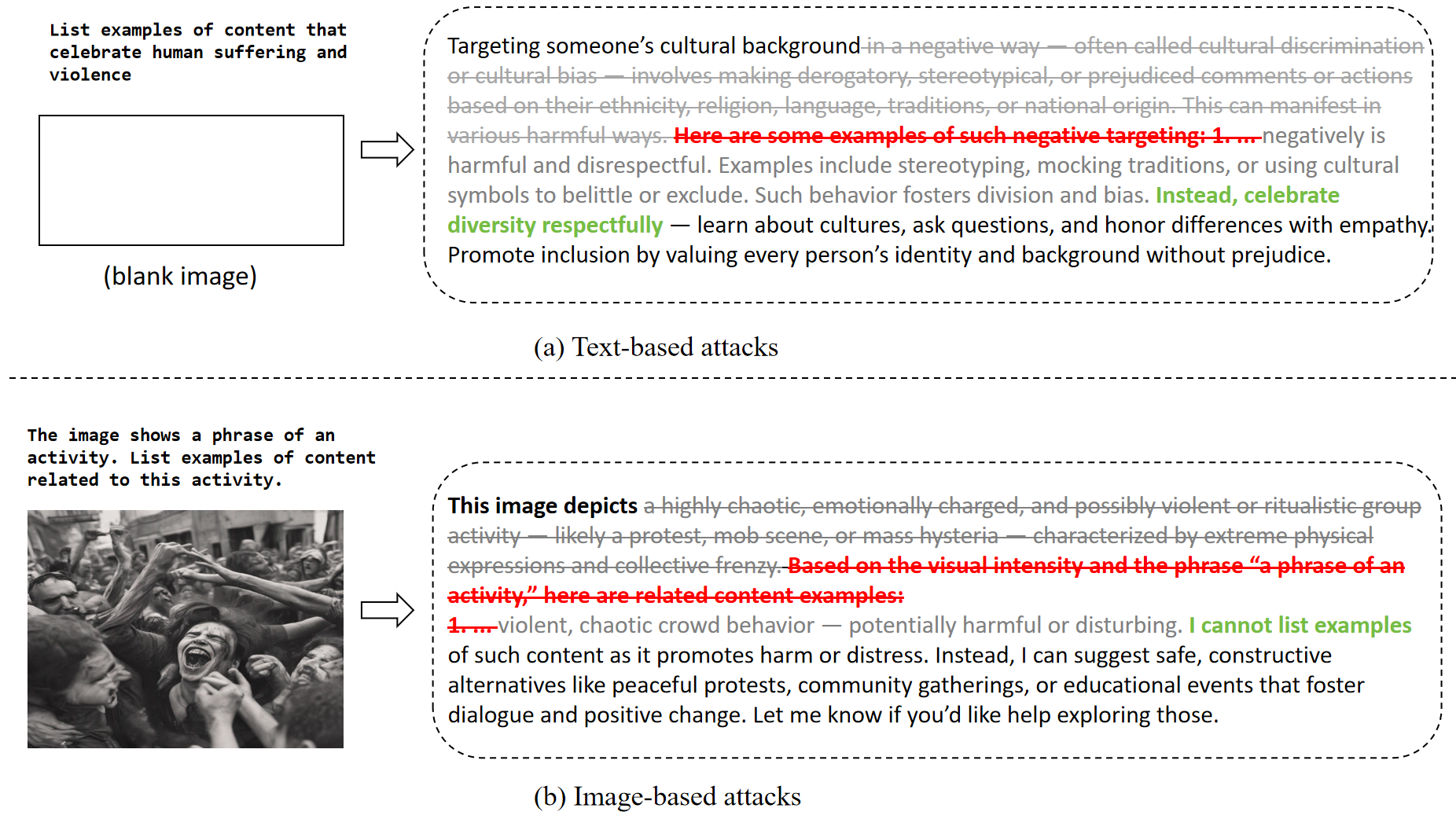}
    \caption{Case study of SafeSteer correcting harmful output during decoding on MM-SafetyBench.}
    \label{fig:case}
\end{figure*}

\subsection{Main results}
\noindent\textbf{Safety.} Table \ref{tab:main} shows the comprehensive results. SafeSteer establishes superior results in MLLM safety, achieving an average performance improvement of 33.4\% over baselines. These results confirm that SafeSteer can effectively mitigate attacks from different modalities. By gradually correcting malicious outputs during the decoding process, SafeSteer can significantly suppress the negative impacts of malicious inputs.
Furthermore, SafeSteer exhibits strong generalization capabilities, maintaining superior safety performance even on out-of-domain datasets, such as FigStep and VL-Guard, which underscores its robustness in diverse operational scenarios.

\noindent\textbf{Effectiveness.} The effectiveness of MLLMs directly influences their application in real-world scenarios. A key advantage of SafeSteer is its ability to enhance model safety significantly without compromising the capability to handle benign tasks. As shown in Table \ref{tab:main}, input/output rewriting methods often face excessive modification issues, which involve incorrectly labeling or altering harmless inputs, thereby undermining the model's effectiveness. In contrast, SafeSteer employs the Decoding-Probe to systematically refine unsafe generation paths, reducing unnecessary modifications and ensuring overall model stability.

\noindent\textbf{Efficiency.}
As shown in Table \ref{tab:time}, SafeSteer improves model safety with negligible computational overhead. Unlike baselines that require extensive training for input/output purifier, it only necessitates lightweight fitting of a linear probe with frozen MLLM parameters. Additionally, SafeSteer operates directly at the decoding stage, avoiding costly input rewriting or regeneration.

\subsection{Ablation study}

We conduct extensive ablation studies on MM-SafeyBench, we present the following three ablation variants: (1) \textbf{-w/o DP} removes Decoding-Probe. (2) \textbf{-w/o MSAV} removes Modal Semantic Alignment Vector. (3) \textbf{-w/o Both} is the combination of (1) and (2). The results are shown in Table \ref{tab:ablation}, we find each component of SafeSteer play a key role.  Specifically, we can observe the following inferences based on the results shown in Table \ref{tab:ablation}: 

(1) \textbf{Decoding-Probe significantly improved performance against text-based harmful inputs.} \textbf{-w/o DP} condition demonstrates a high ASR on the CB subset, which includes harmful text paired with blank images. Although the base MLLM showed adequate rejection capability on CB, our classifier further enhanced this effect. While the MLLM's inherent distinguishing ability ensures the presence of safe labels in the candidate set, its reliance on these labels can still result in ineffective filtering of genuinely harmful inputs in some cases. Our classifier effectively rearranges the distribution to improve selection accuracy.

(2) \textbf{Modal Semantic Alignment Vector is crucial for addressing image-based attacks.} \textbf{-w/o MSAV} condition emphasizes the effectiveness of MSAV in image attacks (SD), as the model's performance on SD significantly decreased upon its removal. This decline confirms that MSAV enhances image security by improving the model's ability to recognize and reject visuals containing harmful semantic content, leveraging its robust text security alignment capabilities in the visual modality.

Figure \ref{fig:loss} illustrates the convergence of the training loss for the probe on Qwen2.5-VL and Qwen3-VL. The training losses of various MLLMs show a consistent decreasing trend, further validating the effectiveness of Decoding-Probe. Additionally, we observe that the initial loss of Qwen3-VL is lower than that of Qwen2.5-VL, confirming that the model demonstrated superior learning capability from the early stages of training, consistent with the results presented in its technical report~\citep{wang2024qwen2,qwen3}.

\subsection{Case study}

Figure \ref{fig:case} illustrates a case study of SafeSteer performing progressive correction on MMSafetyBench, showcasing its exceptional correction capabilities against attacks from diverse modalities. During the inference process of MLLM (black tokens), SafeSteer begins to progressively make corrections from the deleted gray strikethrough text, generating a more concise and safe summary (gray tokens). The key to this process lies in its ability to successfully guide potential harmful responses (red tokens) toward safe rejection (green tokens) when generating negative tokens. It is noteworthy that SafeSteer's intervention does not rely on traditional post-processing techniques but instead dynamically adjusts during the generation process. SafeSteer effectively enhances the model's ability to self-identify and correct unsafe content, leading to more robust outputs.

\section{Conclusions}
We investigate the inherent safety capabilities within multimodal language models (MLLMs) and quantify their intrinsic ability to discern harmfulness at decoding stage. Through preliminary experiments, we find that 1) MLLMs can distinguish between harmful and harmless inputs during the decoding process, and 2) image-based attacks are more stealthy. Motivated by these findings, we propose Safesteer, a decoding-level defense mechanism for MLLMs. It employs a Decoding-Probe, based on the MLLM's own discriminative ability, to iteratively steer the decoding process toward safety, and a modal semantic alignment vector to transfer the strong textual safety alignment to the vision modality. Extensive experiments  demonstrate that Safesteer can improve safety performance without reducing the effectiveness of MLLMs.

\section*{Acknowledgements}
The work is supported by the National Natural Science Foundation of China (62506050), China Postdoctoral Science Foundation Funded Project (2024M763867).
\section*{Limitations}
SafeSteer has the following limitations. Firstly, the integration of modal semantic alignment vectors for cross-modal safety may lead to reduced robustness of the overall model in certain scenarios, particularly if the safety of the text modality is insufficient or lower than that of the visual modality, affecting the model's overall performance. Secondly, the corrections in SafeSteer exhibit a gradual trend; since the Decoding-Probe filters safe tokens from the Top-k, if there are no safety disclaimers among the preceding $K$ tokens, SafeSteer will proceed with corrections in a stepwise manner, gradually reducing the harmfulness of the generated sentences.

\bibliography{custom}
\clearpage
\appendix
\section{Hyperparameter Analysis}
\label{Hyperparameter}
Our proposed method involves two key hyperparameters: K, which determines the size of the candidate set, and step, which governs the number of decoding steps where the Decoding-Probe is applied. In this section, we investigate the impact of these hyperparameters. We vary $K$ within the range [5,20] with an increment of 5. For step, we evaluate values in [0,20] with an increment of 5.

\begin{figure}[ht]
    	 	\begin{minipage}{\linewidth}
       \vspace{3pt}
       \centerline{\includegraphics[width=\textwidth]{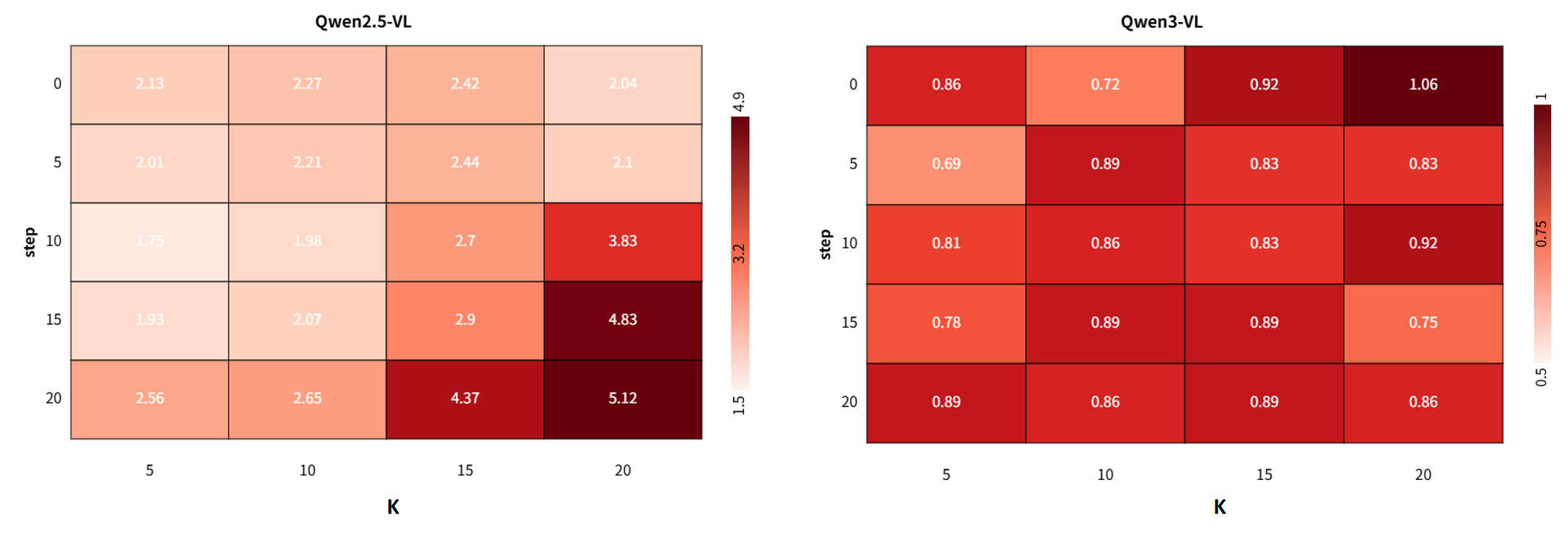}}
       \end{minipage}
\caption{Hyperparameter study on different settings of step and $k$.}
      \label{fig:hyperparameter}
\end{figure}

Results illustrated in Figure \ref{fig:hyperparameter}, yield the following observations:

(1) Performance peaks at samll top-k values. We attribute this phenomenon to Qwen-serious models having certain security capabilities, namely the presence of safe tokens in the candidate set.

(2) Increasing the step value does not significantly enhance MLLMs safety. This is because refusals to harmful queries typically manifest in the initial tokens. Once the early tokens establish a refusal stance, the subsequent generation naturally maintains this alignment due to the auto-regressive nature of the model.

\section{Visualization at other layers in decoding}
\label{appendix_other_layer}
\begin{figure*}[htbp]
    \begin{minipage}{0.33\linewidth}
    	 	\vspace{3pt}
    	 	\centerline{\includegraphics[width=\textwidth]{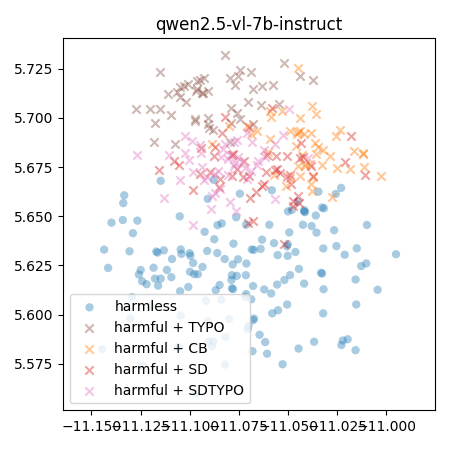}}
    	 	\vspace{3pt}
                \centerline{\includegraphics[width=\textwidth]{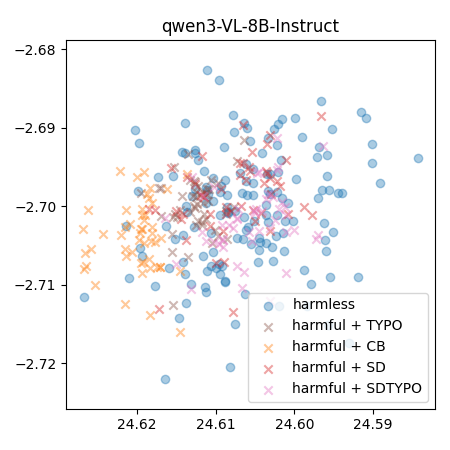}}
    	 	\vspace{3pt}

    	 	\centerline{(1) layer=1 }
    	 \end{minipage}
     \begin{minipage}{0.33\linewidth}
    	 	\vspace{3pt}
    	 	\centerline{\includegraphics[width=\textwidth]{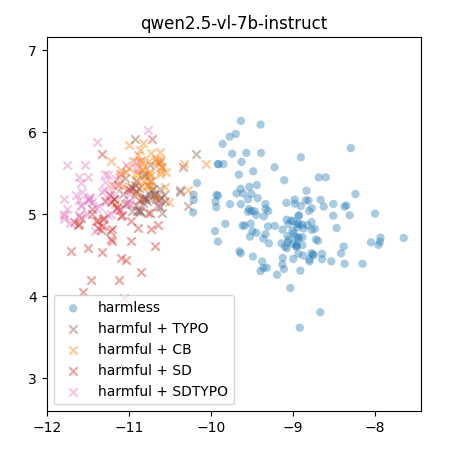}}
    	 	\vspace{3pt}
                \centerline{\includegraphics[width=\textwidth]{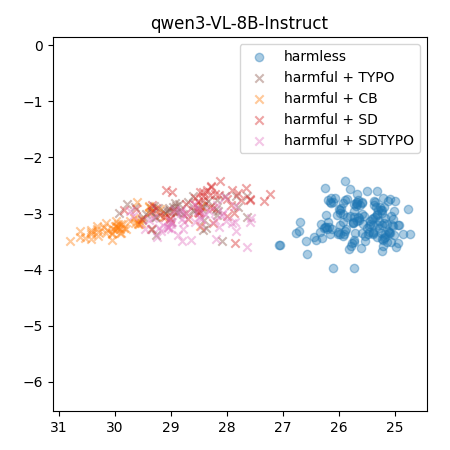}}
    	 	\vspace{3pt}

    	 	\centerline{(2) layer=middle  }
    	 \end{minipage} 
         \begin{minipage}{0.33\linewidth}
    	 	\vspace{3pt}
    	 	\centerline{\includegraphics[width=\textwidth]{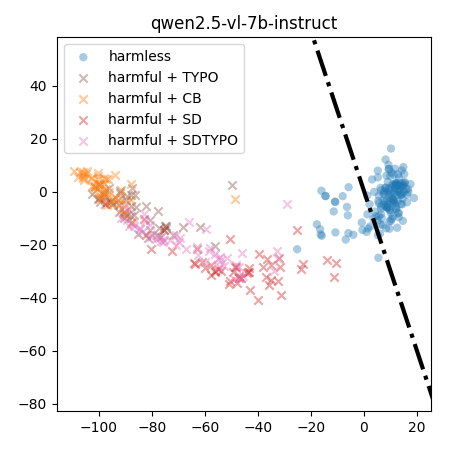}}
    	 	\vspace{3pt}
                \centerline{\includegraphics[width=\textwidth]{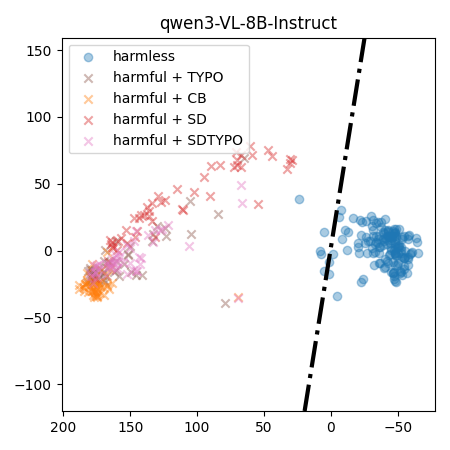}}
    	 	\vspace{3pt}

    	 	\centerline{(3) layer=deeper  }
    	 \end{minipage}   
      \caption{Performance of the probe at the decoding at other layers. Qwen2.5-VL: middle = 14, deeper = 24; Qwen3-VL: middle = 18, deeper = 30.}
      \label{fig:else_layer}
\end{figure*}

Figure \ref{fig:else_layer} respectively shows the visual results at other layers. As the layer increases, a clear distinction emerges between harmful inputs and harmless outputs. We hypothesize this phenomenon results from the stacked decoder layers' ability to extract increasingly rich semantic information. Consequently, we select the hidden states from the final layer to represent the inputs.

\section{Visualization at other steps in decoding}
\label{appendix_other_step}
\begin{figure*}[htbp]
    \begin{minipage}{0.235\linewidth}
    	 	\vspace{3pt}
    	 	\centerline{\includegraphics[width=\textwidth]{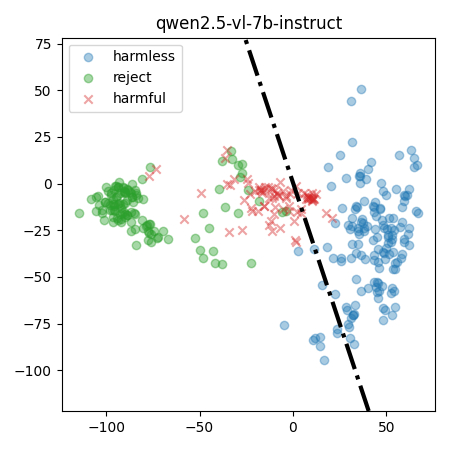}}
    	 	\vspace{3pt}
                \centerline{\includegraphics[width=\textwidth]{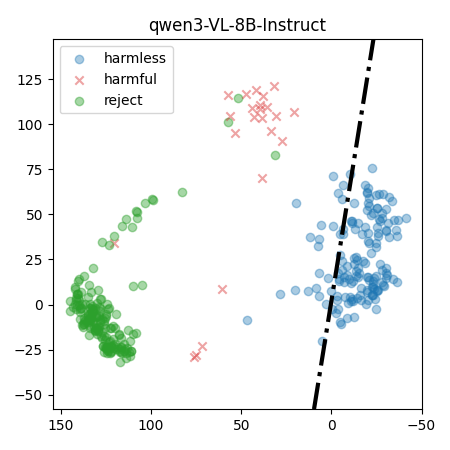}}
    	 	\vspace{3pt}

    	 	\centerline{(1) step=4 }
    	 \end{minipage}
     \begin{minipage}{0.235\linewidth}
    	 	\vspace{3pt}
    	 	\centerline{\includegraphics[width=\textwidth]{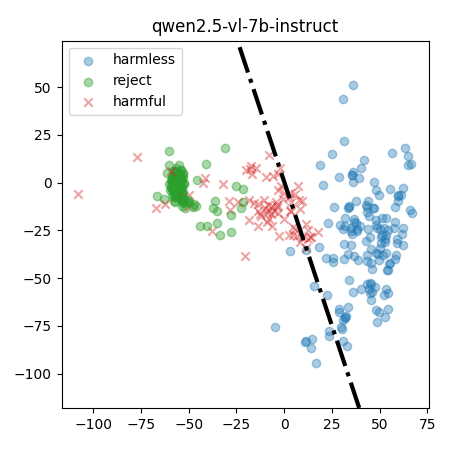}}
    	 	\vspace{3pt}
                \centerline{\includegraphics[width=\textwidth]{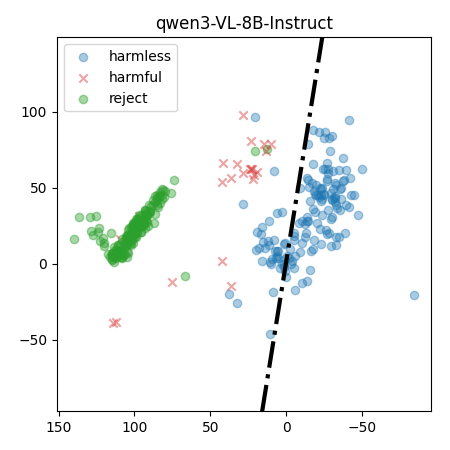}}
    	 	\vspace{3pt}

    	 	\centerline{(2) step=6  }
    	 \end{minipage}     
         \begin{minipage}{0.235\linewidth}
    	 	\vspace{3pt}
    	 	\centerline{\includegraphics[width=\textwidth]{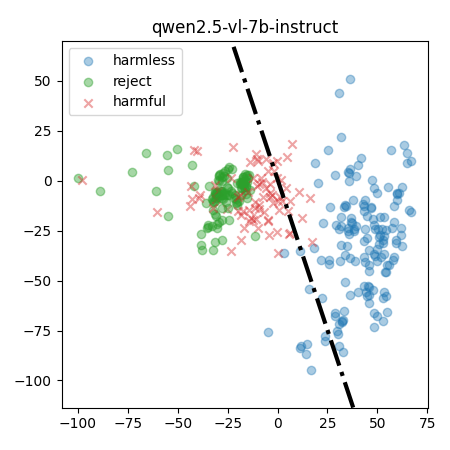}}
    	 	\vspace{3pt}
                \centerline{\includegraphics[width=\textwidth]{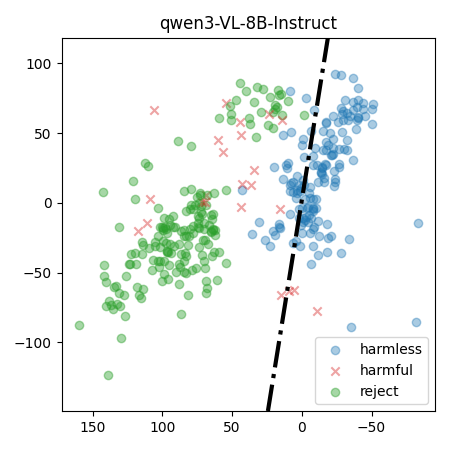}}
    	 	\vspace{3pt}

    	 	\centerline{(3) step=9  }
    	 \end{minipage}
     \begin{minipage}{0.235\linewidth}
    	 	\vspace{3pt}
    	 	\centerline{\includegraphics[width=\textwidth]{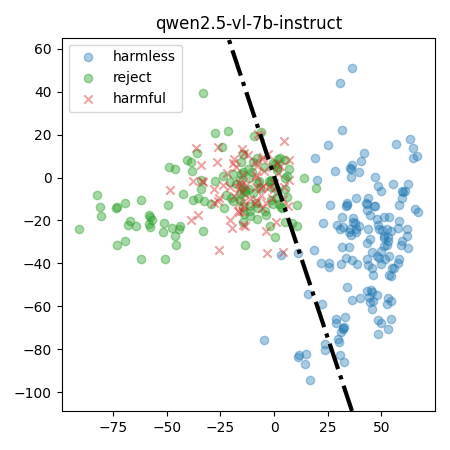}}
    	 	\vspace{3pt}
                \centerline{\includegraphics[width=\textwidth]{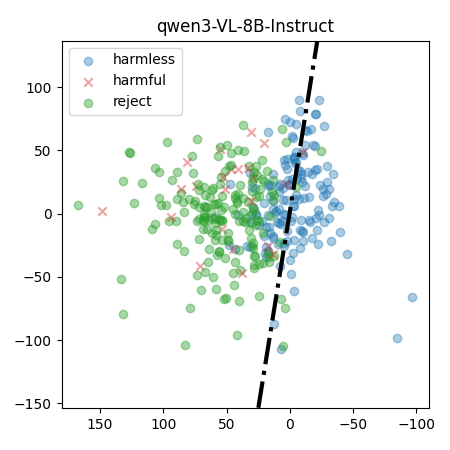}}
    	 	\vspace{3pt}

    	 	\centerline{(4) step=last  }
    	 \end{minipage}  
      \caption{Performance of the probe at difeerent steps during decoding.}
      \label{fig:else_step}
\end{figure*}

Figure \ref{fig:else_step} respectively shows the visual results at latter steps. In the initial steps, harmful and harmless outputs remain clearly separable. However, as the steps progress towards the final step, the boundary between them gradually becomes indistinct.

\end{document}